\def\year{2019}\relax
\documentclass[letterpaper]{article} 
\usepackage{aaai19}  
\usepackage{times}  
\usepackage{helvet}  
\usepackage{courier}  
\usepackage{url}  
\usepackage{graphicx}  
\usepackage{amsmath,amssymb,amsfonts}
\usepackage{bm}
\usepackage{multirow}
\begin{document}
%
\title{DeepCF: A Unified Framework of Representation Learning and Matching Function Learning in Recommender System}
\author{Zhi-Hong Deng$^{1}$, Ling Huang$^{1}$, Chang-Dong Wang$^{1}$, Jian-Huang Lai$^{1}$, Philip S. Yu$^{2,3}$\\
$^{1}$School of Data and Computer Science, Sun Yat-sen University, Guangzhou, China\\
$^{2}$Department of Computer Science, University of Illinois at Chicago, Chicago, USA\\
$^{3}$Institute for Data Science, Tsinghua University, Beijing, China\\
dengzhh7@mail2.sysu.edu.cn, huanglinghl@hotmail.com, changdongwang@hotmail.com\\
stsljh@mail.sysu.edu.cn, psyu@uic.edu
}
\maketitle
\begin{abstract}
In general, recommendation can be viewed as a matching problem, i.e., match proper items for proper users. However, due to the huge semantic gap between users and items, it's almost impossible to directly match users and items in their initial representation spaces. To solve this problem, many methods have been studied, which can be generally categorized into two types, i.e., representation learning-based CF methods and matching function learning-based CF methods. Representation learning-based CF methods try to map users and items into a common representation space. In this case, the higher similarity between a user and an item in that space implies they match better. Matching function learning-based CF methods try to directly learn the complex matching function that maps user-item pairs to matching scores. Although both methods are well developed, they suffer from two fundamental flaws, i.e., the limited expressiveness of dot product and the weakness in capturing low-rank relations respectively. To this end, we propose a general framework named DeepCF, short for Deep Collaborative Filtering, to combine the strengths of the two types of methods and overcome such flaws. Extensive experiments on four publicly available datasets demonstrate the effectiveness of the proposed DeepCF framework.
\end{abstract}

\section{Introduction}
\label{sec:introduction}

Over the past decades, with the fast development of web-based service platforms such as e-commerce platforms and news/music/movie platforms, recommender systems have been extensively studied and widely deployed in many different scenarios to alleviate the information overload problem~\cite{Hu_Item:18,Srivastava2018}. Due to the distinguishing capability of utilizing collective wisdoms and experiences, Collaborative Filtering (CF) algorithms, especially Matrix Factorization (MF) algorithms, have been widely used to build recommender systems~\cite{Wang_Serendipitous:18,Zhao_LSCD:18,Hu_ItemRec:17}.

Matrix factorization assumes some relationship can be established between users and items through some latent factors. By mapping users and items into a common representation space in which they can be compared directly, the similarity between them can be further used to estimate how well they match. In this case, the model learns low-dimensional dense representation for user and item, and then adopts dot product as matching function to calculate matching score. Since Deep Neural Networks (DNNs) are extremely good at representation learning, deep learning methods have been widely explored and have shown promising results in various areas such as computer vision and natural language processing~\cite{he2016deep,serban2016building}. In the past few years, there are also many works adopting DNNs to introduce auxiliary data such as images, text descriptions and demographic information, to improve the representation learning process. What's more, in vanilla matrix factorization, the mapping between the original representation space and the latent space is assumed to be linear, which can not be always guaranteed. To better learn the complex mapping between these two spaces, Xue et al.~\cite{xue2017deep} proposed a Deep Matrix Factorization (DMF), which uses a two pathway neural network architecture to replace the linear embedding operation used in vanilla matrix factorization. However, when it comes to the matching score prediction, matrix factorization methods still resort to dot product which simply combines latent factors linearly and seriously limits the expressiveness of the model.

In addition to learning better representation for users and items, DNNs are very suitable to learn the complex matching function since they are capable of approximating any continuous function~\cite{hornik1989multilayer}. For example, He et al.~\cite{he2017neural} proposed NeuMF under the Neural Collaborative Filtering (NCF) framework which takes the concatenation of user embedding and item embedding as the input of a Multi-Layer Perceptron (MLP) model to make prediction. The high capacity and nonlinearity of DNNs is used to learn the complex mapping relation between user-item representation and matching score. In this case, MLP is used to replace dot product used in traditional matrix factorization methods. However, as revealed in~\cite{beutel2018latent}, MLP is very inefficient in catching low-rank relations. In fact, using dot product to estimate matching score in traditional matrix factorization methods is to artificially limit the model to learn similarity --- a low-rank relation that is thought to be positively related to matching score according to human experience. Although using MLP to learn the matching function directly endows the model with a great flexibility, without introducing human experience, the learning process may be inefficient. This is also why NeuMF needs to incorporate MLP with a shallow matrix factorization model.

According to the above discussion, we can see that there are two types of methods for implementing collaborative filtering~\cite{xu2018deep}. One is based on representation learning and the other one is based on matching function learning. To overcome the shortages of these two types of methods and further improve the performance of CF methods, we incorporate them under the proposed DeepCF framework. In particular, we first use these two types of CF methods to obtain different representations for the input user-item pair. Since these two types of methods have different advantages and learn the representation from different perspectives, a stronger and more robust joint representation for the user-item pair can be obtained by concatenating their learned representations. To calculate the matching score, we then pass this joint representation into a fully connected layer which enables the model to assign different weights on the features. Besides, since the quantity of implicit data far outweighs the quantity of explicit data in real world, designing recommendation algorithms that can work with implicit feedback data is extremely important and has been one of the hot research topics in recommender system. As a result, we focus on implicit feedback in this paper.

The main contributions of this work are as follows.
\begin{itemize}
\item We point out the significance of incorporating collaborative filtering methods based on representation learning and matching function learning, and present a general Deep Collaborative Filtering (DeepCF) framework. The proposed framework abandons the traditional {Deep+Shallow} pattern and adopts deep models only to implement collaborative filtering with implicit feedback.
\item We propose a novel model named Collaborative Filtering Network (CFNet) based on the vanilla MLP model under the DeepCF framework, which has great flexibility to learn the complex matching function while being efficient to learn low-rank relations between users and items.
\item We conduct extensive experiments on four real-world datasets to demonstrate the effectiveness and rationality of the proposed DeepCF framework.
\end{itemize}

\section{Related Work}
\label{sec:relatedwork}

\subsection{Collaborative Filtering with Implicit Data} 
Since most of users would not tend to rate items, it's often difficult to collect explicit feedback. As a result, the quantity of implicit data, such as a click, view, collect, or purchase, far outweighs the quantity of explicit data, such as a rating or a like. In this case, it's very important to design recommendation algorithms that can work with implicit feedback data~\cite{oard1998implicit}. The well-known ALS model~\cite{hu2008collaborative} and SVD++ model~\cite{koren2008factorization} are the early exploration that studied collaborative filtering on datasets
with implicit feedback. Both of the two models factorize the binary interaction matrix and assume user dislike unselected items, i.e., assign 0 for unselected items in the binary interaction matrix. Several works have been done to further improve collaborative filtering with implicit data by assuming user prefer the selected items than the unselected items~\cite{rendle2009bpr,mnih2012learning,he2016vbpr}.

\subsection{Collaborative Filtering based on Representation Learning}
Since Simon Funk proposed Funk-SVD~\cite{funk2006svd} in the famous Netflix Prize competition, matrix factorization for collaborative filtering has been widely studied and constantly developed over the past ten years~\cite{salakhutdinov2008bayesian,koren2009matrix,koren2009collaborative,ma2013experimental,hu2014your}. Although these works tried to improve matrix factorization from different ways, e.g., introducing time, social information, text description, and location, their main idea is still mapping user and item into a common representation space where they can be compared directly. Recently, deep learning methods have shown promising results in various areas such as computer vision, speech recognition and natural language processing. There are also some works proposed to use DNNs for collaborative filtering based on representation learning. AutoRec~\cite{sedhain2015autorec} is the first model attempting to learn user representation and item representation by using autoencoder to reconstruct the input ratings. Collaborative Denoising Auto-Encoders (CDAE)~\cite{wu2016collaborative} further improved it by inputting both ratings and IDs. On the other hand, DMF~\cite{xue2017deep} uses a two pathway neural network architecture to factorize rating matrix and learn representations. Overall, representation learning-based methods learn representation in different ways and can flexibly incorporate with auxiliary data such as images, text descriptions, demographic information and so on. However, they still resort to the dot product or cosine similarity when predicting matching score.

\subsection{Collaborative Filtering based on Matching Function Learning} 
{NeuMF}~\cite{he2017neural} is a recently proposed framework that replaces the dot product used in vanilla MF with a neural network to learn the matching function. To offset the weakness of MLP in capturing low-rank relations, {NeuMF} unifies MF and MLP in one model. {NNCF}~\cite{bai2017neural} is a variant of {NeuMF} that takes user neighbors and item neighbors as inputs. {ConvNCF}~\cite{he2018outer} uses an outer product
operation to replace concatenation used in {NeuMF} so that the model can better learn pairwise correlations between embedding dimensions. Other than {NeuMF}, there are also many other works attempting to learn the matching function directly by making full use of auxiliary data. For example, {Wide\&Deep}~\cite{cheng2016wide} adapts LR and MLP to learn the matching function from input continuous features and categorical features of user and item. {DeepFM}~\cite{guo2017deepfm} replaces LR with Factorization Machines (FM) to avoid manual feature engineering. {NFM}~\cite{he2017nfm} proposed to use a bi-interation pooling layer to learn feature crosses. What's more, tree-based models are also studied and proven to be effective~\cite{zhao2017gb,zhu2017deep,wang2018tem}. In this paper, we focus on pure collaborative filtering without using auxiliary data. In this case, we mainly discuss {NeuMF} and compare it with the proposed DeepCF framework. 

According to the above discussion, both representation learning-based and matching function learning-based collaborative filtering methods have been broadly studied and proven to be effective. Despite their strengths, both of the two types of methods have weaknesses, i.e., the limit expressiveness of dot product and the weakness in capturing low-rank relations. To our best knowledge, so far there is no work to point out the significance of combining the strengths of the two types of collaborative filtering methods to overcome these weaknesses. In this paper, we present a general framework that ensembles these two types of methods to endow the model with a great flexibility of learning the matching function while maintaining the ability to learn low-rank relations efficiently.

\section{Preliminaries}
\label{sec:preliminaries}

\subsection{Problem Statement}
Suppose there are $M$ users and $N$ items in the system, following~\cite{wu2016collaborative,he2017neural}, we construct the user-item interaction matrix $\bm{\mathrm{Y}} \in \mathbb{R}^{M \times N}$ from users' implicit feedback as follows,
\begin{align}\label{eq:interaction_matrix}
y_{ui} = 
\begin{cases}
1,& \text{if interaction (user } u \text{, item } i\text{) is observed;}\\
0,& \text{otherwise.}
\end{cases}
\end{align}

Comparing with explicit feedback, implicit feedback has two major problems. First, unlike ratings, an observed interaction ($y_{ui} = 1$) can only reflects users' preference indirectly, i.e., it can not tell how much exactly a user likes an item. Second, an unobserved interaction ($y_{ui} = 0$) does not mean user $u$ does not like item $i$. In fact, user $u$ may have never seen item $i$ since there are too many items in a system. These two problems pose huge challenges in learning from implicit data, especially the second one. 

To perform collaborative filtering on implicit data which lacks real negative feedback is also known as the One-Class Collaborative Filtering (OCCF) problem~\cite{pan2008one}. In general, there are two ways to tackle this problem, one is to treat all unobserved interactions as weak negative instances~\cite{hu2008collaborative,pan2008one} and the other is to sample some negative instances from unobserved interactions~\cite{pan2008one,wu2016collaborative,he2017neural}. In this paper, we prefer the second method, i.e., uniformly sample negative instances from unobserved interactions.

The recommendation problem with explicit feedback is usually formulated as a rating prediction problem which estimates the missing values in rating matrix $\bm{\mathrm{R}}$. The predicted scores are then used for ranking items and finally the top-ranking items are recommended to users. Similarly, to tackle the recommendation problem with implicit feedback, we can formulate it as an interaction prediction problem which estimates the missing values in interaction matrix $\bm{\mathrm{Y}}$, i.e., estimates whether the unobserved interactions would happen or not. However, unlike explicit feedback, implicit feedback is discrete and binary. Solving the above binary classification problem can not help us to further rank and recommend items. One feasible solution is to employ a probabilistic treatment for interaction matrix $\bm{\mathrm{Y}}$. We can assume $y_{ui}$ obeys a Bernoulli distribution:
\begin{align}
\label{eq:bernoulli_distribution}
\nonumber P(y_{ui} = k \vert p_{ui}) =&
\begin{cases}
1-p_{ui},& k=0;\\
p_{ui},& k=1
\end{cases}\\
=& p_{ui}^k(1-p_{ui})^{1-k},
\end{align}
where $p_{ui}$ is the probability of $y_{ui}$ being equal to 1. What's more, $p_{ui}$ can be further interpreted as the probability that user $u$ is matched by item $i$. In this case, a value of 1 for $p_{ui}$ indicates that item $i$ perfectly matches user $u$ and a value of 0 indicates that user $u$ and item $i$ do not match at all. Rather than modeling $y_{ui}$ which is discrete and binary, our method models $p_{ui}$ instead. In this manner, we transform the binary classification problem, i.e., the interaction prediction problem, to a matching score prediction problem.

\subsection{Learning the Model}

A model-based method generally assumes that data can be generated by an underlying model as $\hat{y}_{ui} = f(u, i \vert \Theta)$, where $\hat{y}_{ui}$ denotes the prediction of $y_{ui}$, i.e., the predicted probability that user $u$ is matched by item $i$, $\Theta$ denotes model parameters, and $f$ denotes the function that maps model parameters to the predicted score. In this manner, we need to figure out two key questions, i.e., how to define function $f$ and how to estimate parameters $\Theta$. We will answer the first question in the next section.

For the second question, most of the existing works generally estimate parameters through optimizing an objective function. Two types of objective functions are commonly used in recommender system, namely, {point-wise loss}~\cite{hu2008collaborative,he2016fast} and {pair-wise loss}~\cite{rendle2009bpr,mnih2012learning,he2016vbpr}. In this paper, we explore the point-wise loss only and leave the pair-wise loss in our future work. Point-wise loss has been widely studied in collaborative filtering with explicit feedback under the regression framework~\cite{funk2006svd,salakhutdinov2008bayesian}. The most commonly used point-wise loss is the {squared loss} (SE). However, the squared loss is not suitable for implicit feedback because it's derived by assuming the error between the given rating $r_{ui}$ and the predicted rating $\hat{r}_{ui}$ obeys a normal distribution, which does not hold in the implicit feedback scenario since $y_{ui}$ is discrete and binary. As aforementioned in \textbf{Problem Statement}, we assume $y_{ui}$ obeys a Bernoulli distribution, i.e., $y_{ui} \sim  Bern(p_{ui})$. By replacing $p_{ui}$ with $\hat{y}_{ui}$ in Equation~\ref{eq:bernoulli_distribution}, we can define the likelihood function as
\begin{align}\label{eq:likelihood_function}
\begin{split}
L(\Theta)
& = \prod_{(u,i) \in \mathcal{Y}^+ \cup \mathcal{Y}^-} P(y_{ui} \vert \Theta )\\
& = \prod_{(u,i) \in \mathcal{Y}^+ \cup \mathcal{Y}^-} \hat{y}_{ui}^{y_{ui}}(1-\hat{y}_{ui})^{1-y_{ui}},
\end{split}
\end{align}
where $\mathcal{Y}^+$ denotes all the observed interactions in $\bm{\mathrm{Y}}$ and $\mathcal{Y}^-$ denotes the sampled unobserved interactions, i.e., the negative instances. Furthermore, taking the negative logarithm of the likelihood (NLL), we obtain
\begin{align}\label{eq:NLL}
\ell_{BCE}
= -\sum_{(u,i) \in \mathcal{Y}^+ \cup \mathcal{Y}^-} y_{ui}\log\hat{y}_{ui} + (1-y_{ui})\log(1-\hat{y}_{ui}).
\end{align}
Based on all the above assumptions and formulations, we finally obtain an  objective function which is suitable for learning from implicit feedback data, i.e., the binary cross-entropy loss function.

To sum up, the recommendation problem with implicit feedback can be formulated as an interaction prediction problem. To endow algorithm with the ability to rank items, we employ a probabilistic treatment for interaction matrix $\bm{\mathrm{Y}}$ so that $y_{ui}$ is assumed to obey a Bernoulli distribution. Instead of modeling $y_{ui}$, we model $p_{ui}$ which is the probability of $y_{ui}$ being equal to 1. Since $p_{ui}$ can also be interpreted as the probability that user $u$ is matched by item $i$, the interaction prediction problem can be transformed to a matching score prediction problem. In this manner, using maximum likelihood estimation to estimate model parameters $\Theta$ is equivalent to minimizing the binary cross-entropy between $y_{ui}$ and $\hat{y}_{ui}$. 

\section{The Proposed Framework}
\label{sec:theproposedframework}

In this section, we first introduce the general processes of representation learning-based CF methods and matching function learning-based CF methods. Then we elaborate these two types of methods and their MLP implements we used in this paper. Finally we illustrate how to fuse these two methods in the proposed DeepCF framework and how to learn the final model.

\subsection{The General Process}
\begin{figure}[!t]
    \centering
    \includegraphics[width=1.0\linewidth]{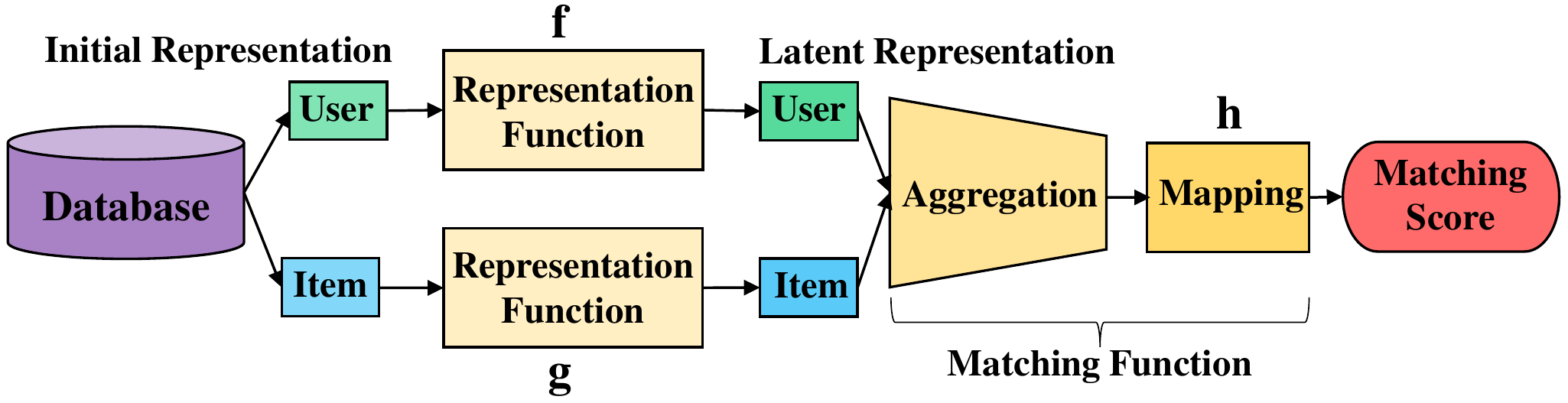}
    \caption{The general process for representation learning-based CF methods and matching function learning-based CF methods.}
    \label{fig:flowchart}
\end{figure}

The processes for representation learning-based CF methods and matching function learning-based CF methods can be concluded as the workflow shown in \figurename~\ref{fig:flowchart}. Both of the two types of methods start from extracting data from database. IDs, historical behaviors and other auxiliary data can all be used to construct the initial representations of user $u$ and item $i$, which are denoted by $\bm{\mathrm{v}}_u^U$ and $\bm{\mathrm{v}}_i^I$ respectively. The CF models then calculate $\bm{\mathrm{p}}_u = f(\bm{\mathrm{v}}_u^U)$ and $\bm{\mathrm{q}}_i = g(\bm{\mathrm{v}}_i^I)$, i.e., the latent representations for user $u$ and item $i$. Next, a non-parametric operation is performed on $\bm{\mathrm{p}}_u$ and $\bm{\mathrm{q}}_i$ to aggregate the latent representations. Finally, mapping function $h(\cdot)$ is used to calculate the matching score $\hat{y}_{ui}$. Notice that the last two steps are referred to as the matching function.

\subsection{Representation Learning}
For representation learning-based CF methods, the model focuses more on learning representation function and the matching function is usually assumed to be simple and non-parametric, e.g., dot product or cosine similarity. In this manner, the model is supposed to learn to map users and items into a common space where they can be directly compared. For example, taking one-hot IDs as inputs, the vanilla MF~\cite{funk2006svd} adopts linear embedding function as function $f(\cdot)$ and function $g(\cdot)$ to learn the latent representations. The latent representations $\bm{\mathrm{p}}_u$ and $\bm{\mathrm{q}}_i$ are then aggregated by the dot product function to calculate the matching score. In this case, mapping function $h(\cdot)$ is assumed to be the identity function. For another example, taking ratings as inputs, DMF~\cite{xue2017deep} adopts MLP as function $f(\cdot)$ and function $g(\cdot)$ to learn better latent representation by making full use of the non-linearity and high capacity characteristics of neural networks. The cosine similarity between $\bm{\mathrm{p}}_u$ and $\bm{\mathrm{q}}_i$ is then calculated and used as matching score.

In this paper, we focus on implicit data only so no auxiliary data are used. The user-item interaction matrix $\bm{\mathrm{Y}}$ is taken as input, i.e., user $u$ is represented by the corresponding row $\bm{y}_{u*}$ in $\bm{\mathrm{Y}}$ and item $i$ is represented by the corresponding column $\bm{y}_{*i}$ in $\bm{\mathrm{Y}}$. In this paper, we adopt MLP to learn latent representations for users and items. Therefore, the representation learning part for users can be defined as:
\begin{align}\label{eq:rlrl}
\begin{split}
\bm{\mathrm{a}}_0
& = \bm{\mathrm{W}}_0^T\bm{y}_{u*}\\
\bm{\mathrm{a}}_1
& = a(\bm{\mathrm{W}}_1^T\bm{\mathrm{a}}_0 + \bm{\mathrm{b}}_1)\\
& \cdot\cdot\cdot\cdot\cdot\cdot\\
\bm{\mathrm{p}}_u = \bm{\mathrm{a}}_\mathrm{X}
& = a(\bm{\mathrm{W}}_\mathrm{X}^T\bm{\mathrm{a}}_{\mathrm{X}-1} + \bm{\mathrm{b}}_\mathrm{X}),
\end{split}
\end{align}
where $\bm{\mathrm{W}}_x$, $\bm{\mathrm{b}}_x$, and $\bm{\mathrm{a}}_x$ denote the weight matrix, bias vector and activation for the $x$-th layer's perceptron respectively. $a(\cdot)$ is the activation function and we use $ReLU$ function in this paper. The latent representation $\bm{\mathrm{q}}_i$ for item $i$ is calculated in the same manner. Different from the existing representation learning-based CF methods, the matching function part is defined as:
\begin{align}\label{eq:rlml}
\hat{y}_{ui} = \sigma(\bm{\mathrm{W}}_{out}^T(\bm{\mathrm{p}}_u \odot \bm{\mathrm{q}}_i)),
\end{align}
where $\bm{\mathrm{W}}_{out}$ and $\sigma(\cdot)$ denote the weight matrix and the sigmoid function respectively. By substituting the non-parametric {dot product or} cosine similarity with element-wise product and a parametric neural network layer, our model still focuses on catching low-rank relations between users and items but is more expressive since the importance of latent dimensions can be different and the mapping can be non-linear. 

In summary, the representation learning component used in this paper is implemented by Equation~\ref{eq:rlrl} and  Equation~\ref{eq:rlml}, which is called CFNet-rl.

\subsection{Matching Function Learning}
Although matching function learning-based CF methods focus more on matching function learning. The representation learning part is still necessary since $\bm{\mathrm{v}}_u^U$ and $\bm{\mathrm{v}}_i^I$ are usually extremely sparse and have high dimension, making it difficult for the model to directly learn the matching function. Therefore, matching function learning-based CF methods usually use a linear embedding layer to learn latent representations for users and items. With the dense low-dimensional latent representations, the model is able to learn the matching function more efficiently.

In this paper, we adopt MLP to learn the matching function. Instead of IDs, we take the interaction matrix $\bm{\mathrm{Y}}$ as input. Therefore, the matching function learning component can be formalized as:
\begin{align}\label{eq:ml}
\begin{split}
\bm{\mathrm{p}}_u
& = \bm{\mathrm{P}}^T\bm{y}_{u*}\\
\bm{\mathrm{q}}_i
& = \bm{\mathrm{Q}}^T\bm{y}_{*i}\\
\bm{\mathrm{a}}_0
& = \begin{bmatrix}\bm{\mathrm{p}}_u\\\bm{\mathrm{q}}_i\end{bmatrix}\\
\bm{\mathrm{a}}_1
& = a(\bm{\mathrm{W}}_1^T\bm{\mathrm{a}}_0 + \bm{\mathrm{b}}_1)\\
& \cdot\cdot\cdot\cdot\cdot\cdot\\
\bm{\mathrm{a}}_\mathrm{Y}
& = a(\bm{\mathrm{W}}_\mathrm{Y}^T\bm{\mathrm{a}}_{\mathrm{Y}-1} + \bm{\mathrm{b}}_\mathrm{Y})\\
\hat{y}_{ui} 
& = \sigma(\bm{\mathrm{W}}_{out}^T\bm{\mathrm{a}}_\mathrm{Y}),
\end{split}
\end{align}
where $\bm{\mathrm{P}}$ and $\bm{\mathrm{Q}}$ are the parameter matrices of the linear embedding layers. The meanings of other notions are the same as CFNet-rl. In this manner, the representation learning functions $f(\cdot)$ and $g(\cdot)$ are implemented by the linear embedding layers. The latent representations $\bm{\mathrm{p}}_u$ and $\bm{\mathrm{q}}_i$ are then aggregated by a simple concatenation operation. Finally, MLP is used as the mapping function $h(\cdot)$ to calculate the matching score $\hat{y}_{ui}$. Notice that although concatenation is the simplest aggregation operation, it maintains maximally the information passed from the previous layer and allows to make full use of the flexibility of the MLP model.

In summary, the matching function learning component used in this paper is implemented by Equation~\ref{eq:ml}, which is called CFNet-ml.

\begin{figure}[!t]
    \centering
    \includegraphics[width=1.0\linewidth]{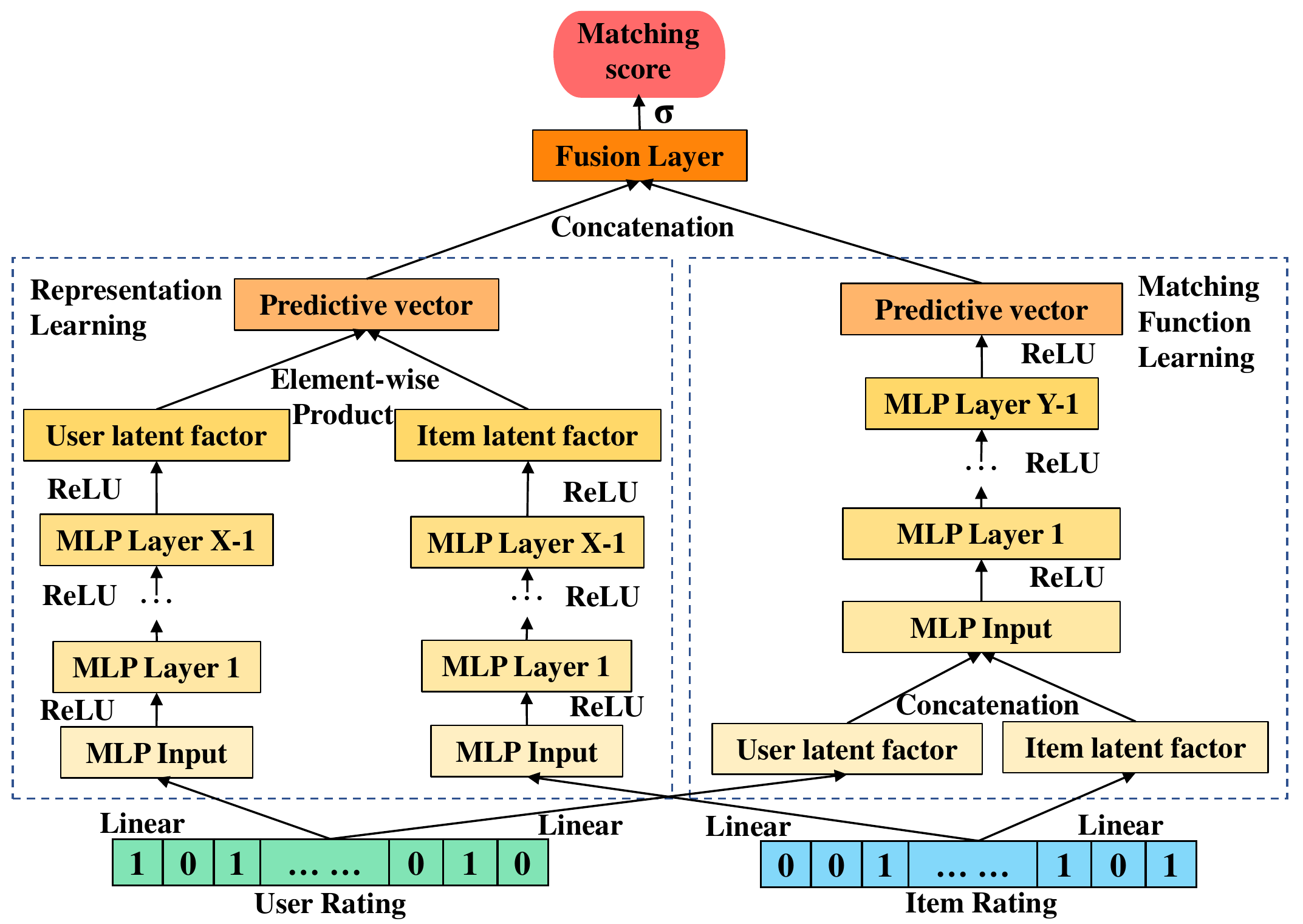}
    \caption{The architecture of CFNet.}
    \label{fig:architecture}
\end{figure}

\subsection{Fusion and Learning}

\subsubsection{Fusion} In the previous two subsections, we have presented the MLP implementations for the two types of methods, i.e., the CFNet-rl model and the CFNet-ml model. To incorporate these two models, we need to design a strategy to fuse them so that they can enhance each other. One of the most common fusing strategies is to concatenate the learned representations to obtain a joint representation and then feed it into a fully connected layer. In our case, for CFNet-rl, the matching function shown in Equation~\ref{eq:rlml} can be split into two steps. The model first calculates the element-wise product for user latent factor and item latent factor, and then sums it up with different weights. The product vector obtained in the first step is called the predictive vector in this paper. For CFNet-ml, the last layer of MLP is called the predictive vector as well. In both cases, the predictive vectors can be viewed as the representation for the corresponding user-item pair. Since the two types of CF methods have different advantages and learn the predictive vectors from different perspectives, the concatenation of the two predictive vectors will result in a stronger and more robust joint representation for the user-item pair. What's more, the consequent fully connected layer enables the model to assign different weights on the features contained in the joint representation. Suppose we denote the predictive vectors of the representation learning component and the matching function learning component as $\bm{\mathrm{a}}_\mathrm{Y}^{rl}$ and $\bm{\mathrm{a}}_\mathrm{Y}^{ml}$ respectively, then the output of the fusion model can be defined as:
\begin{align}\label{eq:output}
\hat{y}_{ui} = \sigma(\bm{\mathrm{W}}_{out}^T\begin{bmatrix}\bm{\mathrm{a}}_\mathrm{Y}^{rl}\\\bm{\mathrm{a}}_\mathrm{Y}^{ml}\end{bmatrix}).
\end{align}
Using Equation~\ref{eq:output} to incorporate CFNet-rl and CFNet-ml, we finally obtain the proposed CFNet model. The architecture of CFNet is shown in \figurename~\ref{fig:architecture}.

\begin{table}[!t]
    \caption{Statistics of the Datasets.}
    \label{statisticsOfDatasets}
    \centering
    \begin{tabular}{ccccc}
        \hline
        Statistics&ml-1m&lastfm&AMusic&AToy\\
        \hline
        \hline
        \# of Users&6040&1741&1776&3137\\
        \# of Items&3706&2665&12929&33953\\
        \# of Ratings&1000209&69149&46087&84642\\
        Sparsity&0.9553&0.9851&0.9980&0.9992\\
        \hline
    \end{tabular}
\end{table}

\begin{table*}[!th]
    \caption{Comparison results of different methods in terms of NDCG@10 and HR@10.}
    \label{Comparisons}
    \centering
    \begin{tabular}{cc|cccc|ccc|c}
        \hline
        \multirow{2}*{Datasets}&\multirow{2}*{Measures}&\multicolumn{4}{|c|}{Existing methods}&\multicolumn{3}{|c|}{CFNet}&Improvement of\\
        & &ItemPop&eALS&DMF&NeuMF&CFNet-rl&CFNet-ml&CFNet&CFNet vs. NeuMF\\
        \hline
        \multirow{2}*{ml-1m}&HR&0.4535&0.7018&0.6565&\textbf{0.7210}&0.7127&0.7073&\textbf{0.7253}&0.6\%\\
        &NDCG&0.2542&0.4280&0.3761&\textbf{0.4387}&0.4336&0.4264&\textbf{0.4416}&0.7\%\\
        \hline
        \multirow{2}*{lastfm}&HR&0.6628&0.8265&0.8840&\textbf{0.8868}&0.8840&0.8834&\textbf{0.8995}&1.4\%\\
        &NDCG&0.3862&0.5162&0.5804&\textbf{0.6007}&0.6001&0.5919&\textbf{0.6186}&3.0\%\\
        \hline
        \multirow{2}*{AMusic}&HR&0.2483&0.3711&0.3744&0.3891&0.3947&\textbf{0.4071}&\textbf{0.4116}&5.8\%\\
        &NDCG&0.1304&0.2352&0.2149&0.2391&\textbf{0.2504}&0.2420&\textbf{0.2601}&8.8\%\\
        \hline
        \multirow{2}*{AToy}&HR&0.2840&0.3717&0.3535&0.3650&0.3746&\textbf{0.3931}&\textbf{0.4150}&13.7\%\\
        &NDCG&0.1518&\textbf{0.2434}&0.2016&0.2155&0.2271&0.2293&\textbf{0.2513}&16.6\%\\
        \hline
    \end{tabular}
\end{table*}

\subsubsection{Learning} As discussed in the previous section, the objective function to minimize for the DeepCF framework is the binary cross-entropy function. To optimize the model, we use mini-batch Adam~\cite{kingma2014adam}. The batch size is fixed to 256 and the learning rate is 0.001. The model parameters are randomly initialized with a Gaussian distribution (with a mean of 0 and standard deviation of 0.01) and the negative instances $\mathcal{Y}^-$ are uniformly sampled from unobserved interactions in each iteration.

\subsubsection{Pre-training} According to~\cite{erhan2010does}, the initialization is of significance to the convergence and performance of deep learning model. Using pre-trained models to initialize the ensemble model can significantly increase the convergence speed and improve the final performance. Since CFNet is composed of two components, i.e., CFNet-rl and CFNet-ml, we can pre-train these two components and use them to initialize CFNet. Notice that CFNet-rl and CFNet-ml are trained from scratch using Adam while the CFNet with pre-training is optimized by the vanilla SGD. This is because Adam requires momentum information of the previous updated parameters which is not saved in CFNet with pre-training.

\section{Experiments}
\label{sec:experiments}

In this section, we conduct experiments to demonstrate the effectiveness of the proposed DeepCF framework and its MLP implementation (i.e., the CFNet model). We also verify the utility of pre-training by comparing the CFNet models with and without pre-training. Hype-parameter sensitivity analysis is discussed in the last part of this section. We implement the proposed model based on Keras\footnote{\url{https://github.com/keras-team/keras}} and Tensorflow\footnote{\url{https://github.com/tensorflow/tensorflow}}, which will be released publicly upon acceptance.

\subsection{Experimental Settings}

\subsubsection{Dataset}
We evaluate our models on four public datasets: MovieLens 1M (ml-1m)\footnote{\url{https://grouplens.org/datasets/movielens/}}, LastFM (lastfm)\footnote{\url{http://www.dtic.upf.edu/~ocelma/MusicRecommendationDataset/}}, Amazon music (AMusic) and Amazon toys (AToy)\footnote{\url{http://jmcauley.ucsd.edu/data/amazon/}}. The ml-1m dataset has been preprocessed by the provider. Each user has at least 20 ratings and each item has been rated by at least 5 users. We process the other three datasets in the same way. The statistics of these four datasets are summarized in \tablename~\ref{statisticsOfDatasets}.

\subsubsection{Evaluation Protocols}
Following~\cite{he2017neural}, we adopt the leave-one-out evaluation, i.e., the latest interaction of each user is used for testing. Since ranking all items is time-consuming, we randomly sample 100 unobserved interactions for each user. We then rank the 100 items with the test item according to the prediction. Two widely adopted evaluation measures, namely Hit Ratio (HR) and Normalized Discounted Cumulative Gain (NDCG) are used to evaluate the ranking performance. The ranked list is truncated at 10 for both measures. Intuitively, the HR measures whether the test item is present on the top-10 list or not, and the NDCG measures the ranking quality which assigns higher scores to hit at top position ranks. 

\subsection{Comparison Results}

The comparison methods are as follows.
\begin{itemize}
\item
\textbf{ItemPop} is a non-personalized method that is often used as a benchmark for recommendation tasks. Items are ranked by their popularity, i.e., the number of interactions.
\item
\textbf{eALS}~\cite{he2016fast} is a state-of-the-art MF method. It uses all unobserved interactions as negative instances and weights them non-uniformly by item popularity.  
\item
\textbf{DMF}~\cite{xue2017deep} is a state-of-the-art representation learning-based MF method which performs deep matrix factorization with normalized cross entropy loss as loss function. We ignore the explicit ratings and take the implicit feedback as input.
\item
\textbf{NeuMF}~\cite{he2017neural} is a state-of-the-art matching function learning-based MF method which combines hidden layers of GMF and MLP to learn the interaction function based on cross entropy loss. It is the most related work to the proposed models. Different from our models, it adapts the deep+shallow pattern which has been widely adopted in many works such as~\cite{cheng2016wide,guo2017deepfm}. What's more, NeuMF takes IDs as input and the proposed CFNet takes interaction matrix as input.
\end{itemize}

Since the proposed models focus on modeling the relationship between users and items, we mainly compare with user-item models. The comparison results are listed in \tablename~\ref{Comparisons}. The best and the second best scores are shown in bold. According to the table, we have the following key observations:
\begin{itemize}
    \item CFNet achieves the best performance in general and obtains high improvements over the state-of-the-art methods. Most importantly, such improvement increases along with the increasing of data sparsity, where the datasets are arranged in the order of increasing data sparsity. This justifies the effectiveness of the proposed DeepCF framework that combines representation learning-based CF methods and matching function learning-based CF methods.
    \item The performance of DMF degrades severely when taking implicit feedback as input while the proposed CFNet-rl consistently outperforms it. This indicates that replacing the non-parametric cosine similarity with element-wise product and a parametric neural network layer significantly improves the performance. 
\end{itemize}

\subsection{Impact of Pre-training}
Different from the CFNet with pre-training, we use mini-batch Adam to learn the CFNet without pre-training with random initializations. As shown in \tablename~\ref{Pretraining}, the CFNet with pre-training outperforms the CFNet without pre-training in all cases. This result verifies the utility of the pre-training process which ensures CFNet-rl and CFNet-ml to learn features from different perspectives and therefore allows the model to generate better results.

\subsection{Sensitivity Analysis of Hyperparameters}

\subsubsection{Negative Sampling Ratio} 
\begin{figure}[!t]
    \centering
    \includegraphics[width=1.0\linewidth]{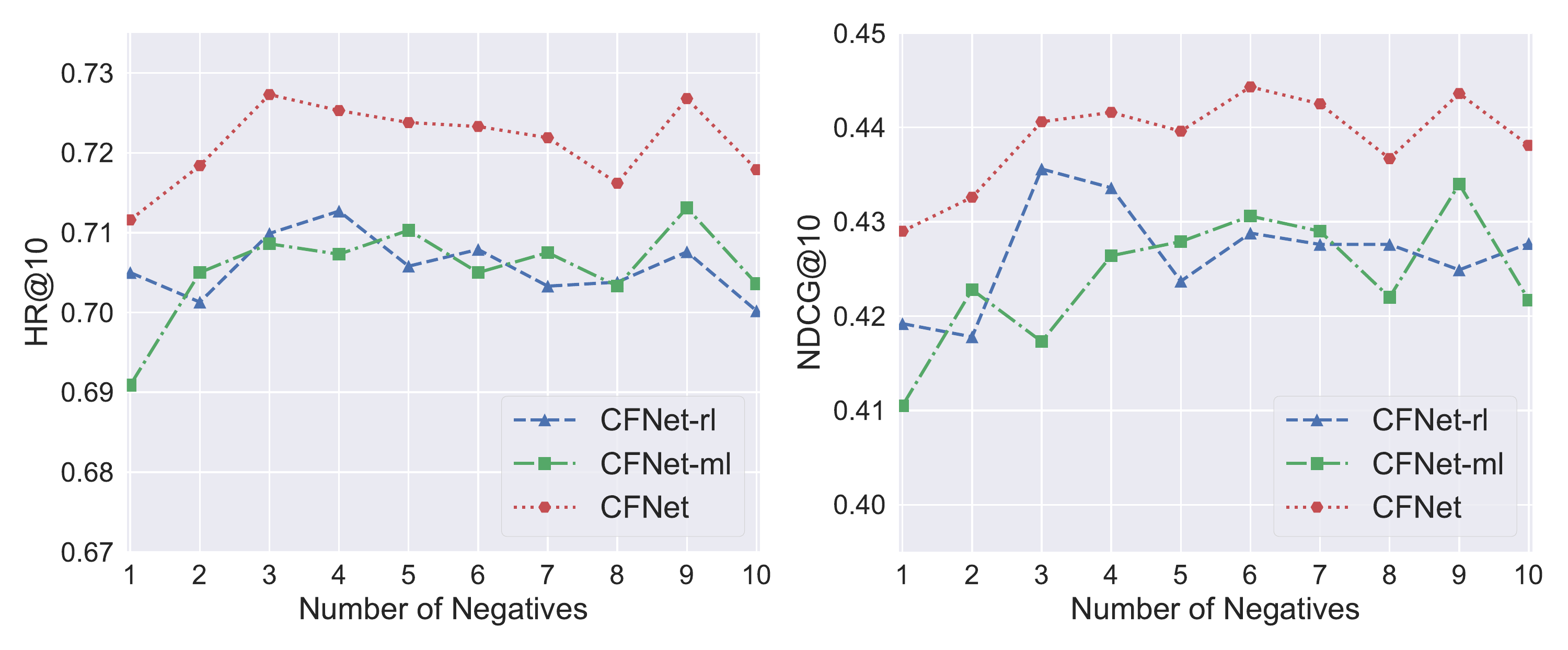}
    \caption{The effect of negative sampling ratio on performance on the ml-1m dataset.}
    \label{fig:parameter_sensitivity_neg}
\end{figure}

To analyze the effect of negative sampling ratio, we test different negative sampling ratio, i.e., the number of negative samples per positive instance, on the ml-1m dataset. From the results shown in \figurename~\ref{fig:parameter_sensitivity_neg}, we can find that sampling merely one or two instances is not enough and sampling more negative instances is helpful. The best HR@10 is obtained when the negative sampling ratio is set to 3 and the best NDCG@10 is obtained when the negative sampling ratio is set to 6. Overall, the optimal sampling ratio is around 3 to 7. Sampling more negative instances not only requires more time to train the model but also degrades the performance, which is consistent with the results shown in~\cite{he2017neural}.

\subsubsection{The Number of Predictive Factors} Another hyper-parameter used in the CFNet model is the number of predictive factors, i.e., the dimensions of $\bm{\mathrm{a}}_\mathrm{Y}^{rl}$ and $\bm{\mathrm{a}}_\mathrm{Y}^{ml}$. As shown in \tablename~\ref{Factors}, the proposed model generates the best performance with 64 predictive factors on most of the datasets except the AMusic dataset. On the Amusic dataset, the best performance is achieved with 16 factors. According to our observation, more predictive factors usually lead to better performances since it endows the model with larger capability and greater ability of representation.

\begin{table}[!t]
    \caption{Performance of CFNet with/without pre-training.}
    \label{Pretraining}
    \centering
    \begin{tabular}{c|cc|cc}
        \hline
        &\multicolumn{2}{|c|}{Without pre-training}&\multicolumn{2}{|c}{With pre-training}\\
        Datasets&HR&NDCG&HR&NDCG\\
        \hline
        ml-1m&0.6962&0.4222&0.7253&0.4416\\
        lastfm&0.8685&0.5920&0.8995&0.6186\\
        AMusic&0.3530&0.2204&0.4116&0.2601\\
        AToy&0.3067&0.1653&0.4150&0.2513\\
        \hline
    \end{tabular}
\end{table}

\begin{table}[!t]
    \caption{Performance of CFNet with different number of predictive factors.}
    \label{Factors}
    \centering
    \begin{tabular}{@{}c@{}c@{}|cccc@{}}
        \hline
        \multirow{2}*{Datasets}&\multirow{2}*{~~Measures~~}&\multicolumn{4}{|c}{Dimensions of predictive vectors}\\
        &&8&16&32&64\\
        \hline
        \multirow{2}*{ml-1m}&HR&0.6820&0.6982&0.7157&0.7253\\
        &NDCG&0.3992&0.4161&0.4351&0.4416\\
        \hline
        \multirow{2}*{lastfm}&HR&0.8840&0.8857&0.8937&0.8995\\
        &NDCG&0.6049&0.6111&0.6143&0.6186\\
        \hline
        \multirow{2}*{AMusic}&HR&0.4003&0.4313&0.4262&0.4116\\
        &NDCG&0.2480&0.2617&0.2661&0.2601\\
        \hline
        \multirow{2}*{AToy}&HR&0.3797&0.3902&0.4026&0.4150\\
        &NDCG&0.2273&0.2331&0.2383&0.2513\\
        \hline
    \end{tabular}
\end{table}

\section{Conclusion and Future Work}
\label{sec:conclusion}

In this work, we have explored the possibility of fusing representation learning-based CF methods and matching function learning-based CF methods. We have devised a general framework DeepCF and proposed its MLP implementation, i.e., CFNet. The DeepCF framework is simple but effective. Although we have implemented the two components with MLP in this paper, different types of representation learning-based methods and matching function learning-based methods can be integrated under the DeepCF framework. This work points out the significance of incorporating these two types of methods, allowing the model to have both great flexibility to learn the complex matching function and high efficiency in learning low-rank relations between users and items. In future work, we will study the following problems. First, auxiliary data can be used to further improve the initial representations of users and items. Richer information usually leads to better performance. Second, except for element-wise product and concatenation, it is also very interesting to explore other aggregation methods. Third,  DeepCF does not only support point-wise loss, using pairwise loss is also a feasible solution for learning the model. Finally, although we use DeepCF to solve the top-N recommendation problem with implicit data, it's also suitable for other data mining tasks that study the relations between two kinds of objects.

\section{ Acknowledgments}
This work was supported by NSFC (61502543, 61876193 and 61672313), Guangdong Natural Science Funds for Distinguished Young Scholar (2016A030306014), Tip-top Scientific and Technical Innovative Youth Talents of Guangdong special support program (2016TQ03X542), National Key Research and Development Program of China (2016YFB1001003), and NSF through grants IIS-1526499, IIS-1763325, and CNS-1626432.

\bibliography{ref}
\bibliographystyle{aaai}
\end{document}